\documentclass[journal]{IEEEtran}
\usepackage{header}
\begin{document}
\title{Dilated Convolutions with Lateral Inhibitions for Semantic Image Segmentation}

\author{Yujiang Wang,  
Mingzhi Dong,
Jie Shen, \IEEEmembership{Member, IEEE,}
Yiming Lin, \IEEEmembership{Graduate Student Member, IEEE,}
and Maja Pantic, \IEEEmembership{Fellow, IEEE}

\thanks{Manuscript received xx xx, xxxx; revised xx xx, xxxx. (Corresponding author: Jie Shen.)}
\thanks{
Yujiang Wang is with Department of Computing, Imperial College London, UK (e-mail: {yujiang.wang14@imperial.ac.uk}).}
\thanks{
Mingzhi Dong is with School of Computer Science, Fudan University, China (e-mail: {mingzhidong@gmail.com}).}
\thanks{
Jie Shen, Yiming Lin, and Maja Pantic are with Department of Computing, Imperial College London, UK (e-mail: {jie.shen07@imperial.ac.uk}; {yiming.lin15@imperial.ac.uk};  {maja.pantic@gmail.com}).
}%
\thanks{Jie Shen is the corresponding author (e-mail: jie.shen07@imperial.ac.uk).}
}

\newcommand\copyrighttext{%
\footnotesize \textcopyright \hspace{0.1mm} This work has been submitted to the IEEE for possible publication. Copyright may be transferred without notice, after which this version may no longer be accessible.
}
\newcommand\copyrightnotice{%
\begin{tikzpicture}[remember picture,overlay]
\node[anchor=south, yshift=4pt] at (current page.south) {\fbox{\parbox{\dimexpr\textwidth-\fboxsep-\fboxrule\relax}{\copyrighttext}}};
\end{tikzpicture}%
}

\maketitle

\copyrightnotice
\begin{abstract}
Dilated convolutions are widely used in deep semantic segmentation models as they can enlarge the filters' receptive field without adding additional weights nor sacrificing spatial resolution. However, as dilated convolutional filters do not possess positional knowledge about the pixels on semantically meaningful contours, they could lead to ambiguous predictions on object boundaries. In addition, although dilating the filter can expand its receptive field, the total number of sampled pixels remains unchanged, which usually comprises a small fraction of the receptive field's total area. Inspired by the Lateral Inhibition (LI) mechanisms in human visual systems, we propose the dilated convolution with lateral inhibitions (LI-Convs) to overcome these limitations. Introducing LI mechanisms improves the convolutional filter's sensitivity to semantic object boundaries. Moreover, since LI-Convs also implicitly take the pixels from the laterally inhibited zones into consideration, they can also extract features at a denser scale. By integrating LI-Convs into the Deeplabv3+ architecture, we propose the Lateral Inhibited Atrous Spatial Pyramid Pooling (LI-ASPP), the Lateral Inhibited MobileNet-V2 (LI-MNV2) and the Lateral Inhibited ResNet (LI-ResNet). Experimental results on three benchmark datasets (PASCAL VOC 2012, CelebAMask-HQ and ADE20K) show that our LI-based segmentation models outperform the baseline on all of them, thus verify the effectiveness and generality of the proposed LI-Convs. 
\footnote{Code is available at: \url{https://github.com/mapleandfire/LI-Convs}}
\end{abstract}

\IEEEpeerreviewmaketitle
\section{Introduction}
Since the introduction of the pioneering Fully Convolutional Networks (FCN) \cite{long2015fully}, deep Convolutional Neural Networks (CNNs) \cite{deeplabv3plus2018, zhao2017pyramid, liu2019auto, zhang2018context,huang2019ccnet} have made impressive progress in semantic image segmentation, a task that performs per-pixel classifications. In deep CNN models, a series of convolutions and spatial poolings are applied to obtain progressively more abstract and more representative feature descriptors with decreasing resolutions. As a consequence, the deepest features can have significantly lower resolution than the original image (e.g. only $1/16$ or $1/32$ of the input size in FCN \cite{long2015fully}), hence it would be difficult to decode these features into the segmentation map at the same size of the input image without losing details. This is a crucial challenge in the semantic segmentation task.

Dilated convolutions \cite{holschneider1990real}, which are first applied to the semantic segmentation task by \cite{yu2015multi, chen2014semantic}, can effectively overcome such difficulties and thus are widely employed in state-of-the-art segmentation methods \cite{liu2019auto,deeplabv3plus2018,yang2018denseaspp,chen2018searching, wang2018understanding}. By inserting zeros (\textit{dilation}) into the convolutional filters, dilated convolutions can observe features from larger areas without increasing the kernel parameters, which is important to the extractions of global semantic features. Besides, it can also produce feature maps that are invariant to input resolutions. In practice, dilated convolutions can be utilised to retain the resolution of the feature maps when encoding representations in the backbone network \cite{yu2017dilated, wang2018understanding}, typically by replacing certain convolutional layers with dilated ones. It can also be employed during the decoding stage to generate more robust semantic labels, e.g. the Atrous Spatial Pyramid Pooling (ASPP) \cite{chen2017rethinking, chen2018deeplab} adopts three parallel dilated convolutions with different dilation rates to aggregate the multi-scale contextual information.

Despite its broad applications, dilated convolutions still have several limitations. The pixels around semantically meaningful contours separate different objects and possess stronger semantic information. In dilated convolution, however, the importance of those pixels are not explicitly accentuated, and therefore such positional significance has to be implicitly learnt. This can leads to ambiguous and misleading boundary labels. Various approaches have been proposed to compensate for such problems and to refine the contour predictions, including the Conditional Random Fields (CRF) \cite{chen2018deeplab,chandra2016fast} and the decoder component in Deeplabv3+ \cite{deeplabv3plus2018}. However, dilated convolution's sensitivity on spotting semantically meaningful edges still leaves room for improvement.

Additionally, although the receptive field of dilated filters is enlarged, the total number of sampled pixels stay the same, which only consist of a small fraction of pixels in the area. The sparse sampling can somehow impair the potentials for dense prediction tasks like semantic segmentation. Similar concerns were addressed in \cite{shen2018gaussian, wang2018understanding, dai2017deformable, yan2014image}, and the proposed improvements include a denser Gaussian sampling process \cite{shen2018gaussian}, a hybrid dilated convolution module \cite{wang2018understanding} and the deformable convolutional filters \cite{dai2017deformable}.

\begin{figure*}[t!]
\centering
\includegraphics[width=0.99\linewidth]{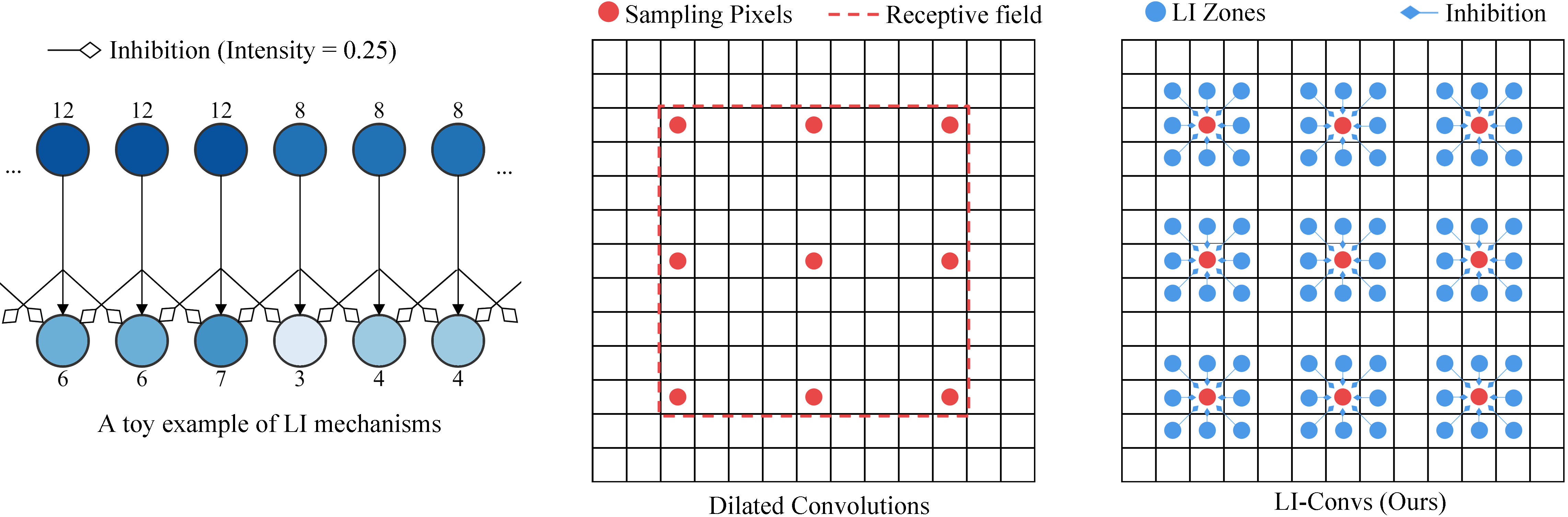}
\caption{\textit{Left}: A toy example to illustrate the lateral inhibition mechanisms where the LI intensity is set to $0.25$. The difference between the two neurons at the centre (representing a semantic contour) becomes more significant after LI. \hspace{1mm} \textit{Middle}: A $3*3$ convolutional filter where $d=4$. The sampled pixels (denoted as red dots) only comprises a small fraction of all pixels in the receptive field. \hspace{1mm} \textit{Right}: An illustration of the proposed LI-Convs with $3*3$ lateral inhibition zones. Each sampled pixel receives inhibition signals from eight neighbours to enhance sensitivity on semantic contours and to extract information at a denser scale.}
\label{fig: method_explain}
\end{figure*}

In this paper, we propose to overcome the drawbacks in dilated convolutions from a biologically-inspired perspective, which is to leverage the Lateral Inhibition (LI) mechanism in the human visual system. Lateral inhibition \cite{hartline1956inhibition, rizzolatti1975inhibition, von2017sensory} is a neurobiological phenomenon that a neuron's excitation to a stimulus can be suppressed by the activation of its surrounding neurons. 
Because of the LI mechanism, our retina cells are sensitive to the spatially varying stimulus such as the semantic borderlines between objects, which is crucial to the inborn segmentation abilities of our eyes. See Fig. \ref{fig: method_explain} (\textit{Left}) for an intuitive illustration of the LI mechanism.

Motivated by such observations, we propose a dilated convolution with lateral inhibitions (LI-Convs) to enhance the convolutional filter's sensitivity to semantic contours. LI-Convs also sample the receptive window in a denser fashion by implicitly making inferences on pixels within the lateral inhibited zones. To evaluate LI-Convs, we follow the Deeplabv3+ \cite{deeplabv3plus2018} segmentation models and present three LI-based variants which are 1). the Lateral Inhibited Atrous Spatial Pyramid Pooling (LI-ASPP) for decoding semantic features, 2). the Lateral Inhibited MobileNet-V2 (LI-MNV2) and 3). the Lateral Inhibited ResNet (LI-ResNet) as the backbone networks for encoding features. The performance of those LI-variants surpasses the baseline on three segmentation benchmark dataset: PASCAL VOC 2012 \cite{everingham2015pascal}, CelebAMask-HQ \cite{CelebAMask-HQ} and ADE20K \cite{zhou2017scene}, which verifies the effectiveness and generality of the proposed LI-Convs.

\section{Related Works}
\textbf{Semantic Image Segmentation} \hspace{0.1cm} 
Fully Convolutional Networks (FCN) \cite{long2015fully} is the pioneering work of using deep models for semantic segmentation. The fully connected layers in deep image classification models are replaced with convolutional ones to produce semantic heat maps for segmentation predictions. The resolution of such heat maps is typically much smaller than that of the input image (e.g. $1/32$), and various works are proposed to compensate the information loss during decoding such features, including the de-convolutional layers \cite{noh2015learning, ronneberger2015u, peng2017large},  the skip-connections of low-level features \cite{badrinarayanan2017segnet, hariharan2015hypercolumns} and dilated convolutions \cite{yu2015multi, chen2017rethinking, yang2018denseaspp, liu2019auto, chen2018searching}. Yu et al. \cite{yu2015multi} stacks dilated convolutional layers with different dilation rates in a cascaded manner, leading to a context module for aggregating the multi-scale contextual information. Deeplabv3 \cite{chen2017rethinking} builds an Atrous Spatial Pyramid Pooling (ASPP) module consisting of three parallel dilated convolutions, one 1*1 convolution and one image-level pooling, and it also employs dilated convolutions in the backbone network.  DenseASPP \cite{yang2018denseaspp} introduces dense connection into the ASPP module to enlarge its receptive fields and to acquire denser feature pyramid, 
while the technique of Neural Architecture Search \cite{zoph2016neural} is utilised by \cite{chen2018searching} to search for an optimal decoding structure of organising dilated convolutions layers. For other segmentation practice \cite{rahman2016optimizing, wang2019face,luo2020shape,wang2019dynamic}, readers are referred to \cite{minaee2020image} for more details.

\textbf{Dilated Convolutions} \hspace{0.1cm}
Dilated convolutions, also known as atrous convolutions, is first introduced by Holschneider et al. \cite{holschneider1990real} in signal analysis and have broad applications such as object detection \cite{li2017pedestrian, nguyen2019lightweight}, lip-reading \cite{xu2020discriminative, martinez2020lipreading, ma2020lip} and optical flow \cite{zhu2018learning, sun2018pwc}. It is first applied to semantic segmentation by authors of \cite{yu2015multi, chen2014semantic} to enlarge filter's receptive fields without sacrificing the spatial resolution. Conditional Random Fields (CRF) are involved in \cite{chandra2016fast, chen2018deeplab} as a post-processing procedure to refine the ambiguous semantic contour predictions. Similar ideas can be found in Deeplabv3+'s decoding module \cite{deeplabv3plus2018} that incorporates low-level backbone features to improve the qualities of contouring pixels. Deformable convolutions \cite{dai2017deformable} introduce the \textit{offsets} into the sampling grids of filters to better model the spatial relationships. Gaussian kernels are adopted by \cite{shen2018gaussian} to obtain pixels at a wider range in dilated convolutions. Wang et al. \cite{wang2018understanding} observe the \textit{gridding} effects brought by the fixed sampling locations in dilated kernels and demonstrate a hybrid dilated convolution with different dilated rates. Different from those approaches, we employ the lateral inhibition (LI) mechanisms \cite{hartline1956inhibition} to enhance the dilated convolutions' sensitivity on semantically meaningful contours and to implicitly sample features in a denser fashion.

\textbf{Lateral Inhibitions} \hspace{0.1cm}
The study on the eyes of horseshoe crab (\textit{Limulus}) performed by Hartline et al. \cite{hartline1956inhibition} reveals the lateral inhibition (LI) effects in visual systems, where the excitation of neighbouring neurons can suppress a cell's response to the stimuli. Although lateral inhibitions are mainly studied in the field of neuroscience \cite{roska2000three, sun2004orientation,rizzolatti1975inhibition}, the computer vision community has also shown interests in this mechanism. The recurrent neural network with lateral inhibitions is studied in \cite{mao2007dynamics} and it is shown that LI can improve the robustness and efficiency of the network. Authors of \cite{fernandes2013lateral} introduce LI into a shallow CNN to improve image classification. Similar ideas can be found in the work for colour video segmentation \cite{fernandez2014color}. Those network architectures are somehow too shallow to be useful for recent methods using deep backbones like MobileNet-V2 (MNV2) \cite{mobilenetv22018} or ResNet \cite{he2016deep}. 
The idea of LI can also be found in the Local Response Normalisation (LRN) proposed by AlexNet \cite{krizhevsky2017imagenet}, yet the inhibitions in LRN come from different channels on the same spatial locations, which might not be suitable for segmentation tasks, also there are no learnable parameters in it.  
Recently, 
authors of \cite{cao2018lateral} employ LI in VGG model \cite{simonyan2014very} to improve the performance on saliency detection. However, none of the previous works has evaluated LI's potentials for semantic segmentation, while their methods of integrating LI do not touch the core mechanisms in deep CNNs such as the convolutional operations. 
In this work, however, lateral inhibitions work closely with the convolutional filters to fundamentally augment the model's segmentation powers.

\section{Dilated Convolutions with Lateral Inhibitions}
\subsection{Definition}
Define $\Psi_k = \mathbb{Z}^2 \cap [-k,k]^2$ where $k \in \mathbb{Z}_{\geq 0}$, and let a discrete function $F: {\Psi_k} \mapsto \mathbb{R}$ represents a convolutional filter of size $(2k+1)^2$. 
Define another discrete function $G : \mathbb{Z}^2 \mapsto \mathbb{R}$ representing features of arbitrary sizes. Let $d$ be the dilation rate, a dilated convolutional operator $*_d$ is written as
\begin{equation}
    \label{eq: dilated_conv_definition}
  (F *_d G)(\mathbf{p}) = \sum_{\mathclap{d\mathbf{m}+\mathbf{n}=\mathbf{p}}} F(\mathbf{m}) G(\mathbf{n})
\end{equation}
where $\mathbf{p},\mathbf{m},\mathbf{n} \in \mathbb{Z}^2$ are 2D spatial indices. Note that $*_d$ turns into a regular convolutional operator when $d=1$, i.e. no dilation is inserted. 

With the introduction of lateral inhibitions (LI), the activation of each sampled pixel, i.e. $G(\mathbf{n})$ in Eq. \ref{eq: dilated_conv_definition}, would be suppressed by its neighbours within a certain range. Let the lateral inhibitions come from a square region of size $(2t+1)^2$ centred on $\mathbf{n}$ where $t \in \mathbb{Z}_{\geq 0}$, and refer this region as the lateral inhibition zone (the LI zone). Define $\Psi_t = \mathbb{Z}^2 \cap [-t, t]^2$ and let $L: {\Psi_t}, \mathbb{Z}^2 \mapsto \mathbb{R}$ be a discrete function describing the spatially-varying inhibition intensities in the LI zones, the amount of the total inhibitions received by a sampled pixel $G(\mathbf{n})$ can be described as $\sum_{\mathbf{u}+\mathbf{v}=\mathbf{n}} L(\mathbf{u},\mathbf{n})G(\mathbf{v})$ where $\mathbf{u},\mathbf{v} \in \mathbb{Z}^2$. Consequently, a dilated convolutional operator with lateral inhibition $\star_d$ (LI-Convs) can be defined as
\begin{equation}
    \label{eq: li_conv_definition}
  (F \star_d G)(\mathbf{p}) =  \sum_{\mathclap{d\mathbf{m}+\mathbf{n}=\mathbf{p}}} F(\mathbf{m}) \phi (G(\mathbf{n}) - \sum_{\mathclap{\mathbf{u}+\mathbf{v}=\mathbf{n}}} L(\mathbf{u},\mathbf{n})G(\mathbf{v}
  ))
\end{equation}
 where $\phi$ represents an activation function like ReLu. The introduced LI terms and non-linearity distinguish LI-Convs in Eq. \ref{eq: li_conv_definition} with Eq. \ref{eq: dilated_conv_definition}. An intuitive comparison between  dilated convolutions and the proposed LI-Convs is shown in
 Fig. \ref{fig: method_explain} (\textit{Middle \& Right}).

We can also \say{dilate} the lateral inhibition zone to efficiently expand its field-of-views, in a similar way to that of dilated convolutions. Consequently, a generalised LI-Convs operator $\star_d^e$ is defined as
\begin{equation}
    \label{eq: general_li_conv_definition}
  (F \star_d^e G)(\mathbf{p}) = \sum_{\mathclap{d\mathbf{m}+\mathbf{n}=\mathbf{p}}} F(\mathbf{m}) \phi (G(\mathbf{n}) - \sum_{\mathclap{e\mathbf{u}+\mathbf{v}=\mathbf{n}}} L(\mathbf{u},\mathbf{n})G(\mathbf{v})
  )
\end{equation}
where $e$ denotes the dilation rate in LI zones. 

Although a wide variety of kernel forms can be taken by the LI intensity descriptor $L$, we opt for an intuitive formulation that is also easy to implement. In particular, $L(\mathbf{u},\mathbf{n})$ in Eq. \ref{eq: general_li_conv_definition} simply takes the production of a differentiable weight $W_L \in [0,1] \cap \mathbb{R}$ and an exponentially decaying factor that is related to the distance between $\mathbf{u}$ and $\mathbf{n}$, which can be described as
\begin{equation}
    \label{eq: l_definition}
L(\mathbf{u},\mathbf{n}) = W_L \exp( \frac{\minus D^2(\mathbf{u},\mathbf{n})}{2\sigma^2})
\end{equation}
where $\sigma$ is a parameter representing the standard deviation, $\exp$ denotes the exponential function and $D(\mathbf{u},\mathbf{n})$ refers to a certain distance measurement between $\mathbf{u}$ and $\mathbf{n}$. Here we employ the Euclidean distance.

\subsection{Implementation of LI-Convs}

\begin{figure*}[ht!]
\centering
\includegraphics[width=0.74\linewidth]{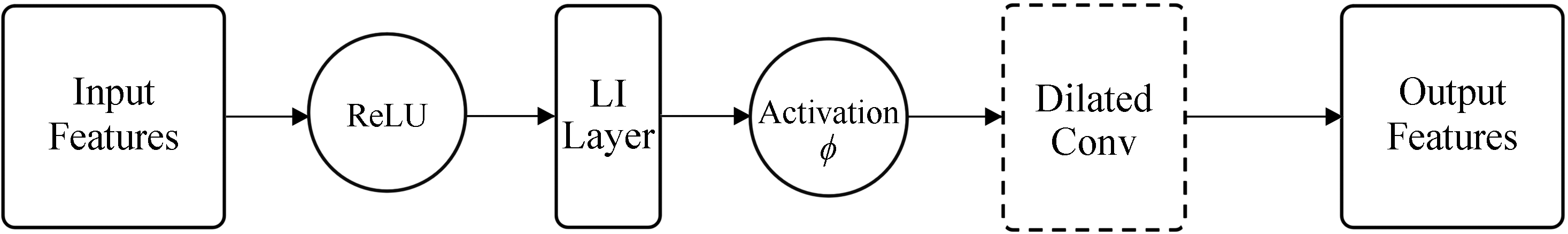}
\caption{The structure of LI-Convs. The lateral inhibitions is first calculated by the LI layer, and the inhibited features are fed into the dilated convolution layer. The dilated convolution part can be any kind of convolution implementations such as the depthwise one \cite{chollet2017xception}. }
\label{fig: li_conv_structures}
\end{figure*}

We take a straight-forward approach to implement the LI-Convs in Eq. \ref{eq: general_li_conv_definition}. We first design a Lateral Inhibition layer (the LI layer) to perform pixel-wise lateral inhibitions, while a dilated convolutional layer is subsequently applied to the inhibited features. The LI layer is essentially a light-weight module that can be flexibly inserted into deep models, while it can be easily implemented as a dilated convolutional layer with specifically shaped filters. In particular, let a discrete function $K : {\Psi_t} \mapsto \mathbb{R}$ represent one such LI filter, $K$ can be described as: 
\begin{align}
\begin{split}
\label{eq: li_layer_kernel}
\displaystyle
K(\mathbf{u}) &= 
    \begin{cases}
        1.0   &   \mathbf{u}=\mathbf{0}. \\
      \minus W_L\exp(\frac{\minus D^2(\mathbf{u},\mathbf{0})}{2\sigma^2})   &  \mathbf{u} \neq \mathbf{0}. \\
    \end{cases} \\
\end{split}
\end{align}
Note that the LI filter $K$ has identical size with the LI zones which is $(2t+1)^2$, and applying $K$ with a stride of 1 can generate pixel-wise inhibited features. We empirically set $\sigma$ in Eq. \ref{eq: li_layer_kernel} to a fixed value during training, thus there is only one weight $W_L$ to learn for each LI filter, which is significantly less than that of regular convolutional filters. In practice, we learn the lateral inhibition weights in a channel-wise manner, i.e. each LI filter learns a separate $W_L$. Therefore, a LI layer will introduce a total of $C$ learnable weights where $C$ is the channel number of the input tensor.

A detailed illustration for the LI-Convs implementations can be found in Fig. \ref{fig: li_conv_structures}. A ReLu activation is first applied to remove negative neuron response. Then a LI layer with filters in Eq. \ref{eq: li_layer_kernel} is employed to extract inhibited features, followed by the activation function $\phi$ in Eq. \ref{eq: general_li_conv_definition}. A dilated convolution layer of arbitrary form such as the depthwise convolution \cite{chollet2017xception} is subsequently employed. 

\subsection{LI-ASPP, LI-MNV2 and LI-ResNet}
\begin{figure}[ht!]
\centering
\includegraphics[width=0.99\linewidth]{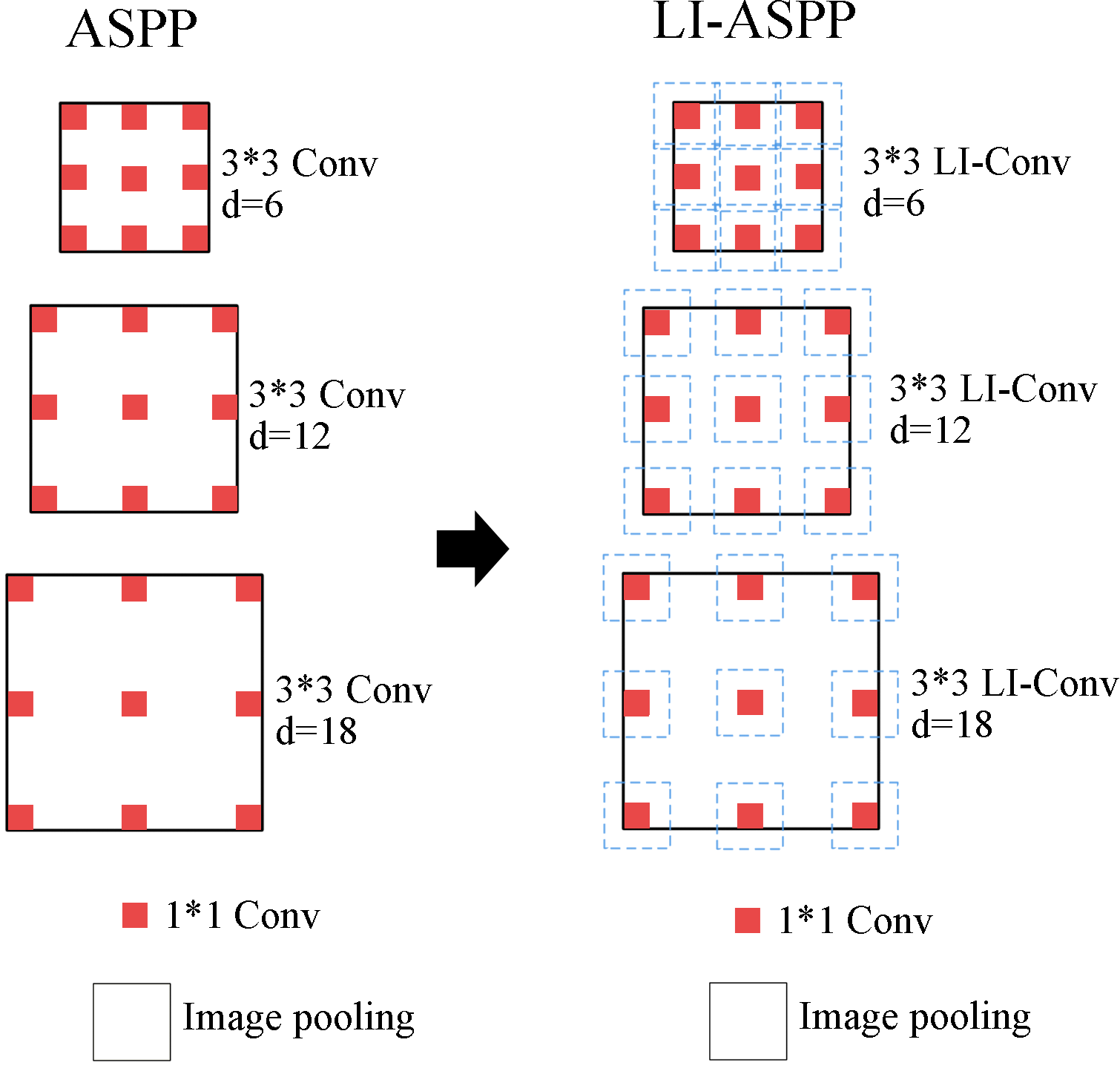}
\caption{
The structures of ASPP and LI-ASPP. ASPP consists of five parallel branches including three dilated convolutions, which are replaced with the proposed LI-Convs in LI-ASPP.}
\label{fig: li_aspp}
\end{figure}
We introduce the proposed LI-Convs into the state-of-the-art segmentation model Deeplabv3+ \cite{deeplabv3plus2018} to evaluate the proposed LI-Convs. 
As shown in Fig. \ref{fig: li_aspp}, we replace the three $3*3$ parallel dilated convolution operations in Atrous Spatial Pyramid Pooling (ASPP) \cite{deeplabv3plus2018} with the proposed LI-Convs, leading to the LI-ASPP model.
Besides, we also investigate the potentials of LI layer in the backbone network such as the MobileNet-V2 (MNV2) \cite{mobilenetv22018} and ResNet \cite{he2016deep}. 

\begin{figure*}[t!]
\centering
\includegraphics[width=0.83\linewidth]{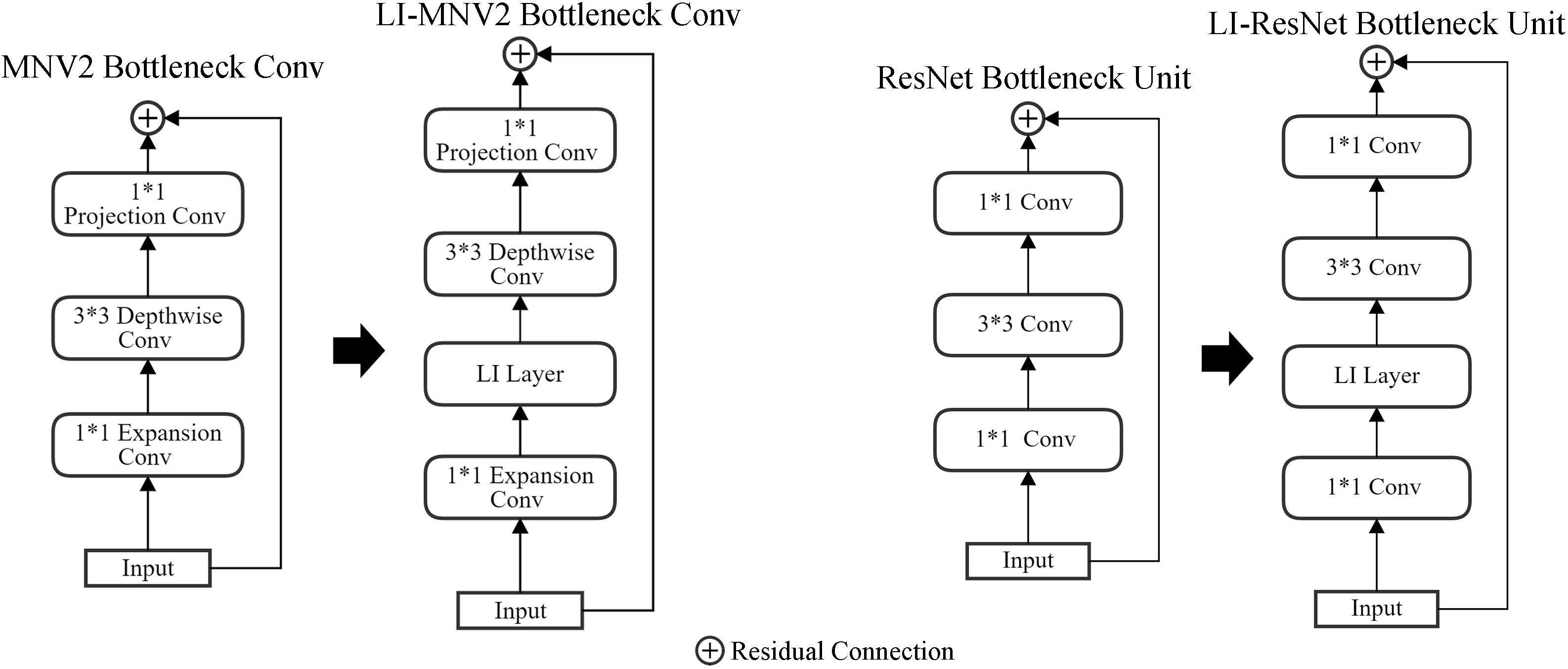}
\caption{\textit{Left}: The structures of the \textit{residual bottleneck} convolution in MobileNet-V2 and the LI bottleneck. The LI layer is inserted between the $1*1$ expansion convolution and $3*3$ depthwise convolution.  \hspace{1mm} \textit{Right}: The architecture of the 3-layer \textit{bottleneck unit} in ResNet and the LI bottleneck unit. The LI layer is inserted between the $1*1$ conv and $3*3$ conv layers.}
\label{fig: li_backbones}
\end{figure*}

As illustrated in Fig. \ref{fig: li_backbones} (\textit{Left}), we insert the LI layer into the \textit{residual bottleneck} (RB) of MobileNet-V2 \cite{mobilenetv22018}, which is between the $1*1$ expansion convolution and $3*3$ depthwise convolution, and we refer the resulting structure as the LI bottleneck layer. 
In the original MNV2 architecture, there are a total of $17$ \textit{residual bottleneck} layers, and we replace the $10^{th}$, $13^{th}$ and $16^{th}$ ($16^{th}$ refers to the second-highest RB layer) \textit{reisudal bottlenecks} with the LI bottlenecks to obtain the LI-MNV2 network. 

Similarly, we modify the \textit{bottleneck unit} (we adopt the one with 3 convolutional layers) in ResNet by inserting a LI layer between the first two weighted layers, as shown in Fig. \ref{fig: li_backbones} (\textit{Right}), and name the new architecture as the LI bottleneck unit. Among those ResNet variants, we select the ResNet-50 architecture in this work and replace its \say{conv5\_3} layer with the LI bottleneck unit to get the LI-ResNet-50 network. 

\section{Experiments}

\subsection{Datasets}
We conduct our experiments on three public benchmark segmentation datasets, which are PASCAL VOC 2012 \cite{everingham2015pascal}, CelebAMask-HQ \cite{CelebAMask-HQ} and ADE20K \cite{zhou2017scene}. There are a total of 21 semantic classes in PASCAL VOC 2012 dataset \cite{everingham2015pascal} which contains 1,464/1,449/1,456 pixel-wise annotated images for train/validation/test. Following \cite{hariharan2011semantic, deeplabv3plus2018}, we use an augmented train set with a total of 10,582 annotated images. CelebAMask-HQ \cite{CelebAMask-HQ} is a large-scale face parsing dataset with 30,000 pixel-wise labelled face images of 19 classes, and they are split into sets with 24,183/2,993/2,824 images for train, validation and test. ADE20K \cite{zhou2017scene} is a benchmark dataset for scene parsing with 20,210/2,000/3,000 pixel-wise labelled images for train/validation/test. It is a quite challenging dataset, as there are a total of 151 classes in this dataset, and the huge variations of image resolutions also increase the difficulties. We utilise the validation set to evaluate performance on PASCAL VOC 2012 and ADE20K datasets, considering that their test sets are not publicly available, while we follow the standard protocol on CelebAMask-HQ dataset and use the test set for evaluation.

\subsection{Experimental Setup}

\textbf{Evaluation metric}  \hspace{0.1cm}
Mean Intersection-over-Union (mIoU) is the most widely used evaluation metric for the segmentation task, and we adopt it to evaluate the quality of model predictions. We also report the model parameters and the FLOPs to provide more comprehensive analyses.

\textbf{Training Settings}  \hspace{0.1cm}
We generally follow the training settings in Deeplabv3+ \cite{deeplabv3plus2018}, while we have also made some modifications to suit our needs. Particularly, we use the ImageNet \cite{russakovsky2015imagenet} checkpoints provided by the authors of MobileNet-V2 \cite{mobilenetv22018} and ResNet \cite{he2016deep} to initialise LI-MNV2 and LI-ResNet-50, respectively, while the weights of LI-ASPP are randomly initialised. Note that we do not use the MS COCO dataset \cite{lin2014microsoft} to pre-train the model. 
During training, we set the image crop size to be $513*513$ for all three datasets, except that we use $257*257$ crop size when evaluating on ADE20K with the MNV2-based backbone.
We train for 120 epochs using a batch size of 16 and Adam \cite{kingma2014adam} is applied to optimise the pixel-wise cross-entropy loss with L2-regularisation. 
The initial learning rate and the epsilon value in Adam optimiser are set to 0.0003 and 0.01, respectively. 
To improve the performance, we additionally fine-tune the LI weights by freezing other weights for another 40 epochs.
The \textit{output stride}, which is defined in \cite{chen2017rethinking} denoting the ratio of original input resolution to the final feature's resolution, is set to be 16 for all datasets. We adopt strategies in \cite{deeplabv3plus2018, chen2017rethinking} to use the BatchNorm layers \cite{ioffe2015batch} and to randomly scale the training data for augmentation. Depthwise convolution \cite{chollet2017xception} is used in the ASPP implementations following \cite{deeplabv3plus2018}. 
During evaluations, we set the \textit{output stride} to be 16 for all datasets and employ a single evaluation scale of 1.0, and all crop sizes are set to $513*513$ except that $257*257$ is utilised for evaluating MNV2-based backbones on ADE20K.

\textbf{LI Layer Settings}  \hspace{0.1cm}
A lateral inhibition layer has several key hyper-parameters that can affect the performance. We fine-tune those parameters on the Pascal Voc 2012 validation set to determine a best-performing combination. Particularly, we set the size of LI zones to be $3*3$, the value for the standard deviation $\sigma$ in Eq. \ref{eq: li_layer_kernel} is selected to be $1.0$, the LI rate $e$ in Eq. \ref{eq: general_li_conv_definition} is set to $1$, and all LI intensities $W_L$ in Eq. \ref{eq: li_layer_kernel} are initialised as $0.0$ such that the training can start smoothly from any pre-trained checkpoints that do not use LI layers.
Moreover, we evaluate different positions of adding LI bottlenecks in MNV2 and ResNet-50 architectures, and a general trend can be observed that adding LI to higher layers can produce better results than to bottom ones. Besides, we select ReLu as the activation $\phi$ in Eq. \ref{eq: general_li_conv_definition}.

\begin{table}[t!]
\centering
\caption{The performance of different LI-Conv's parameters for LI-ASPP on Pascal Voc 2012 validation set. The three numbers in ``LI Zone Sizes'' refers to the sizes of the LI zones (``3'' stands for a $3*3$ LI zone) of the three LI-Conv layers in LI-ASPP, respectively, while ''LI Rates'' indicates the selections of $e$ in Eq. \ref{eq: general_li_conv_definition} for those LI-Conv layers. }
\begin{tabularx}{0.99\columnwidth}{y y k y k k}
\toprule
\textbf{Backbone} & \textbf{Decoding Model} & \textbf{LI Zone Sizes} & \textbf{LI Rates} & \textbf{$W_L$ Init. Range} & \textbf{mIoU} (\%) \\
\midrule
MNV2 & ASPP & - & - & - & 72.19 \\
\midrule
\multirow{5}{*}[0em]{MNV2} & \multirow{5}{*}[0em]{LI-ASPP} & \{3, 3, 3\} & \{1,1,1\} & [0.0,0.0] & \textbf{72.77}  \\
& & \{3, 3, 3\} & \{1,3,5\} & [0.0,0.0] & 72.56 \\
& & \{3, 3, 3\} & \{5,5,5\} & [0.05,0,15] & 72.43 \\
& & \{5, 5, 5\} & \{1,1,1\} & [0.05,0,15] & 72.13 \\
& & \{3, 3, 3\} & \{1,1,1\} & [0.05,0.35] & 72.64 \\
\bottomrule
\end{tabularx}
\label{tab: li_conv_parameters} 
\end{table}

\begin{table}[ht!]
\caption{The performance when adding LI layers to different positions of MNV2 on Pascal Voc 2012 validation set. ``RB'' refers to the \textit{Residual Bottleneck} layer in MNV2 \cite{mobilenetv22018}. }
\centering
\begin{tabularx}{0.99\columnwidth}{k k s k}
\toprule
\textbf{Backbone} & \textbf{Decoding Model} & \textbf{Positions to add LI} & \textbf{mIoU} (\%) \\
\midrule
MNV2 & ASPP & - & 72.19 \\
\midrule
\multirow{4}{*}[0em]{LI-MNV2} & \multirow{4}{*}[0em]{ASPP} & $\{1-6\}^{th}$ RB & 72.07 \\
& & $\{16\}^{th}$ RB & 72.21 \\
& & $\{13,16\}^{th}$ RB & 72.43 \\
& & $\{10,13,16\}^{th}$ RB & \textbf{72.79} \\
\bottomrule 
\end{tabularx}
\label{tab: li_mnv2_positions} 
\end{table}

\begin{table}[ht!]
\caption{The performance of models with ResNet-50-based backbones on Pascal Voc 2012 validation set. We also explore different positions of adding LI layers to ResNet-50. }
\centering
\begin{tabularx}{0.99\columnwidth}{k k s k}
\toprule
\textbf{Backbone} & \textbf{Decoding Model} & \textbf{Positions to add LI} & \textbf{mIoU (\%)} \\
\midrule
ResNet-50 & ASPP & - & 76.22 \\
\midrule
\multirow{3}{*}[0em]{LI-ResNet-50} & \multirow{3}{*}[0em]{ASPP} & conv5\_3 & 76.90 \\
& & conv4\_6, conv5\_3 & 76.53 \\
& & conv3\_4 & 76.21 \\
\midrule 
ResNet-50 & LI-ASPP & Three dilated convs & 76.94 \\
\bottomrule 
\end{tabularx}
\label{tab: li_resnet_performance} 
\end{table}

\textbf{Implementations}  \hspace{0.1cm} 
We implement our method in the Tensorflow framework \cite{abadi2016tensorflow}. For the implementation of the baseline Deeplabv3+ \cite{deeplabv3plus2018} model, we directly use the code provided by authors. 
It takes around one day per GPU (2080TI) to train a model with LI-MNV backbone on Pascal Voc 2012 dataset, and it requires about 2.5/0.6 days to do so on CelebAMask-HQ and ADE20K datasets. For the LI-ResNet-50 backbone, the training will take longer which are approximately 1.5/3.2/4.5 days on Pascal Voc/CelebAMask-HQ/ADE20K using two parallel GPUs.

\subsection{Results}
\begin{table*}[t!]
\caption{Performance of different methods with MNV2-based backbones on the Pascal Voc 2012 and ADE20K (validation set) and on the CelebAMask-HQ (test set). The model parameters and FLOPs (for crop size $513*513$) are also included.}
\centering
\begin{tabularx}{0.84\linewidth}{s j j k y y}
\toprule
\multirow{2}{*}[-0.4em]{\textbf{Method}} & \multicolumn{3}{c}{\textbf{mIoU} (\%)} & \multirow{2}{*}[-0.4em]{\shortstack{\textbf{Parameters} \\ (Kilo)}} & \multirow{2}{*}[-0.4em]{\shortstack{\textbf{FLOPs} \\ (Mega)}}  \\ 
\cmidrule(lr){2-4}
& \textbf{Pascal Voc 2012} & \textbf{CelebAMask-HQ} & \textbf{ADE-20K} & & \\ \midrule
MNV2 + ASPP & 72.19 & 74.73 & 29.97 & 2568.02 & 6479 \\
MNV2 + LI-ASPP & 72.77 & 75.3 & 30.47 & 2568.98 & 6498  \\  
LI-MNV2 + ASPP & 72.79 & 75.46 & 30.59 & 2569.94 & 6517 \\  
LI-MNV2 + LI-ASPP & \textbf{73.14} & \textbf{75.70} & \textbf{30.66} & 2570.52 & 6528 \\ 
\bottomrule
\end{tabularx}
\label{tab: performance_mnv2} 
\end{table*}

\begin{table*}[t!]
\caption{Performance of different methods with ResNet-based backbones on the Pascal Voc 2012 and ADE20K (validation set) and on the CelebAMask-HQ (test set). The performance of disabling/enabling Deeplabv3+ Decoder \cite{deeplabv3plus2018} is reported. 
``RD-ASPP'' refers to replacing each LI layer in \textit{LI-ASPP} with a regular depthwise layer, and similarly for ``RD-ResNet-50''.
The model parameters and FLOPs (for crop size $513*513$) are also included.}
\centering
\begin{tabularx}{0.99\linewidth}{s y k k y i i}
\toprule
\multirow{2}{*}[-0.4em]{\textbf{Method}} & \multirow{2}{*}[-0.4em]{\shortstack{\textbf{Deeplabv3+} \\ \textbf{Decoder} \cite{deeplabv3plus2018}} } & \multicolumn{3}{c}{\textbf{mIoU} (\%)} & \multirow{2}{*}[-0.4em]{\shortstack{\textbf{Parameters} \\ (Kilo)}} & \multirow{2}{*}[-0.4em]{\shortstack{\textbf{FLOPs} \\ (Giga)}}  \\ 
\cmidrule(lr){3-5} 
& & \textbf{Pascal Voc 2012} & \textbf{CelebAMask-HQ} & \textbf{ADE-20K} & & \\ \midrule
ResNet-50 + ASPP & - & 76.22 & 76.03 & 39.14 & 26656 & 87.35 \\
RD-ResNet-50 + RD-ASPP & - & 75.87 & 76.14 & 38.94 & 26716 & 87.49 \\
LI-ResNet-50 + LI-ASPP (Ours) & - & \textbf{77.24} & \textbf{76.62} & \textbf{39.79} & 26663 & 87.48 \\ \midrule
ResNet-50 + ASPP & \checkmark & 77.01 & 78.12 & 40.1 & 26819 & 92.85 \\				
RD-ResNet-50 + RD-ASPP & \checkmark & 76.99 & 78.46 & 40.37 & 26880 & 93.00 \\
LI-ResNet-50 + LI-ASPP (Ours) & \checkmark & \textbf{77.54} & \textbf{79.26} & \textbf{40.92} & 26826 & 92.98 \\
\bottomrule
\end{tabularx}
\label{tab: performance_resnet50} 
\end{table*}

\textbf{LI Parameters} \hspace{0.1cm}  In Table \ref{tab: li_conv_parameters} we demonstrate the performance of different LI parameters for LI-ASPP (with MNV2 as backbone) on Pascal Voc 2012 validation set. In particular, we investigate the performance of varying settings of LI hyper-parameters such as the size of LI Zones, the LI rates $e$ and $W_L$'s initialisation range for the three LI-Convs layers in LI-ASPP. As shown in Table \ref{tab: li_conv_parameters}, most settings can lead to superior performance than the baseline method without any LI-Convs, while using a $3*3$ LI zone and setting $e=1$ can generally yield better performance than other settings like a $5*5$ LI Zone or $e=5$. LI-ASPP achieves the best performance when all $W_L$ is initialised from $0.0$, potentially due to that the zero initialisation can better encourage a smooth learning of LI intensities, and therefore we opt for this setting for all LI layers.

\textbf{Adding LI to MNV2} \hspace{0.1cm} In addition, we evaluate different options of adding LI-Convs in the \textit{Residual Bottleneck} (RB) layers of the MNV2 architecture \cite{mobilenetv22018} on the validation set of Pascal Voc 2012. It can be spotted from Table \ref{tab: li_mnv2_positions} that adding LI mechanisms to the early RB layers (e.g. the earliest six RB layers) cannot promote the accuracy. In contrast, LI-Convs integrated with top layers such as the $\{10,13,16\}^{th}$ RB layers can produce higher mIoUs. This observation is somehow in line with the expectations since the higher-level layers are generally encoding more semantic representations, which can better benefit from the improved sensitivity to semantic contours introduced by LI layers. 

\textbf{Adding LI to ResNet-50} \hspace{0.1cm} Table \ref{tab: li_resnet_performance} demonstrates the results on Pascal Voc 2012 validation set when adding LI layer to different layers of ResNet-50 architecture with ASPP as the decoding model. We can discover that adding LI to earlier layers of ResNet such as the \say{conv3\_4} may not improve the performance, however, top layers like \say{conv4\_6} and \say{conv5\_3} can better benefit from the integration of LI layers. Such observations are consistent with the trend that is found in the LI-MNV2 experiments of Table \ref{tab: li_mnv2_positions}, which is also in accordance with our intuitions for LI layer's effects. A slight difference is that the best result is achieved when LI is added to the \say{conv5\_3} layer other than to both \say{conv4\_6} and \say{conv5\_3} layers. Besides, we report in Table \ref{tab: li_resnet_performance} the performance of LI-ASPP with ResNet-50 as the backbone, which still shows significant improvement over the baseline.

\textbf{Performance Evaluations} \hspace{0.1cm} In Table \ref{tab: performance_mnv2}, we report the evaluation results of different methods with MNV2-based backbones on the three segmentation benchmark datasets. Note that we disable the Deeplabv3+ Decoder \cite{deeplabv3plus2018} in this experiment to ensure a fair and clean comparison. 
Compared with the baseline method which is MNV2+ASPP, i.e. Deeplabv3 \cite{chen2017rethinking}, LI-MNV2 and LI-ASPP both demonstrate superior performance when used solely, while the best mIoUs on three datasets are all achieved by using them together. Particularly, our method (LI-MNV2+LI-ASPP) gains a relative improvement of 1.32\%, 1.30\% and 2.30\% over the baseline (MNV2+ASPP) on Pascal Voc 2012, CelebAMask-HQ and ADE-20K datasets, respectively, which verifies the effectiveness of LI-Convs. The LI-based model's parameters and FLOPs, however, are only slightly increased by 0.097\% and 0.76\% compared with the baseline, which is arguably acceptable considering the accuracy compensations. 

\begin{table*}[t!]
\caption{Performance (measured by mIoU (\%)) of different methods using multi-scale evaluations on Pascal Voc 2012 validation set. ``[1.0]'' refers to using the single evaluation scale, i.e. no multi-scale is utilised.}
\centering
\begin{tabularx}{0.92\linewidth}{s y y k k k}
\toprule
\multirow{2}{*}[-0.8em]{\textbf{Method}} & \multirow{2}{*}[-0.8em]{\shortstack{\textbf{Deeplabv3+} \\ \textbf{Decoder} \cite{deeplabv3plus2018}} } & \multicolumn{4}{c}{\textbf{Evaluation Scales}}  \\ 
\cmidrule(lr){3-6} 
&  & \textbf{[1.0]} & \textbf{[0.5, 1.0, 1.75]} & \shortstack{\textbf{[0.5, 0.75, 1.0,}\\\textbf{1.25, 1.75]}} & \shortstack{[\textbf{0.5, 0.75, 1.0,}\\\textbf{1.25, 1.5, 1.75]}} \\ \midrule
ResNet-50 + ASPP & - & 76.22 & 76.58 & 77.41 & 77.60 \\
LI-ResNet-50 + LI-ASPP & - & 77.24 & 77.93 & 78.37 & 78.58 \\
ResNet-50 + ASPP & \checkmark & 77.01 & 78.33 & 78.71 & 78.72 \\
LI-ResNet-50 + LI-ASPP & \checkmark & 77.54 & 78.66 & 79.05 & 79.19 \\
\bottomrule
\end{tabularx}
\label{tab: performance_multi_eval_scales} 
\end{table*}
\begin{figure*}[ht!]
\centering
\includegraphics[width=0.9\linewidth]{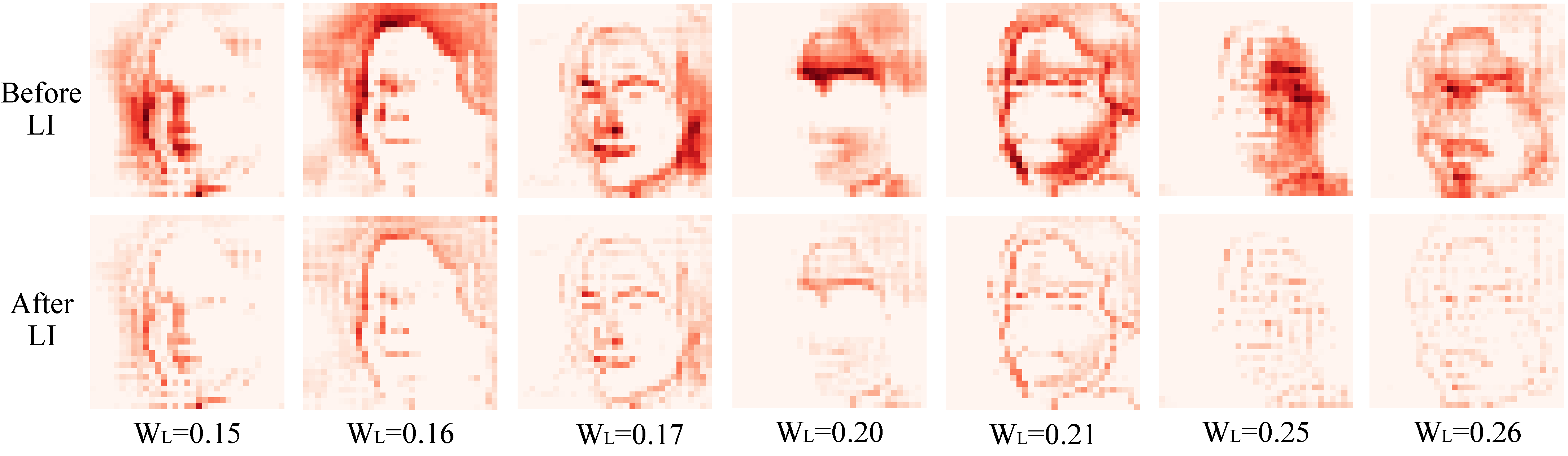}
\caption{Visualisations of the channel-level features before and after LI layers on CelebAMask-HQ. Although the activation is inhibited globally, the feature patterns after LI layer are generally easier to recognise mainly due to the clarifications on semantic contours.}
\label{fig: channel_LI_plots}
\end{figure*}

The evaluation results of ResNet-50-based models are shown in Table \ref{tab: performance_resnet50}, where our methods are additionally compared with 1). Deeplabv3+ Decoder \cite{deeplabv3plus2018} that also aims to refine the semantic contours, and 2). The variants (\say{RD-ASPP} and \say{RD-ResNet-50} in Table \ref{tab: performance_resnet50}) that replaces each LI layer in LI-ASPP and LI-ResNet-50 with a regular depthwise layer of identical kernel size and channels.
We can see from the table that when Deeplabv3+ Decoder is disabled, our method (LI-ResNet-50+LI-ASPP) outperforms the baseline (ResNet-50+ASPP) on all three datasets at the cost of slightly increased parameters and FLOPs, which is consistent with the MNV2-based results in Table \ref{tab: performance_mnv2}. Enabling Deeplabv3+ Decoder introduces mIoU boosts to both our method and the baseline, while our LI models still demonstrate greater improvement over the baseline on all datasets. This indicates that LI-Convs can work closely with Deeplabv3+ Decoder to produce dense predictions with higher-qualities, exhibiting the compatibility and the flexibility of integrating LI-Convs into other deep models.  
Moreover, our LI-based models (LI-ResNet-50+LI-ASPP) with Deeplabv3+ Decoder disabled can achieve similar performance as the baseline (ResNet-50+ASPP) that enables it, while the former model of ours contains 0.58\% fewer parameters and operates at approximately 5.78\% faster speed than the latter one, respectively, which shows LI-Convs's light-weighted features. 
Additionally, when each LI layer in the LI-variants is replaced with a regular depthwise layer of unconstrained kernels, its performance (\say{RD-ASPP+RD-ResNet-50} in Table \ref{tab: performance_resnet50}) have dropped significantly on all three datasets when compared with ours LI-ResNet-50+LI-ASPP, while it also has the largest learnable parameters, therefore demonstrating the advantages of the pre-definfed LI kernels in Eq. \ref{eq: li_layer_kernel}.

\begin{table}[t]
\caption{Training and Inference speed comparison on Pascal Voc 2012. We report the results of both MNV2 and ResNet-50 backbones before and after adding the LI layers. Our proposed LI layer adds very little extra overhead to the training and inference time.}
\label{tab: speed_pascal} 
\centering
\footnotesize
\begin{tabularx}{0.99\linewidth}{j i i i}
\toprule
\multirow{1}{*}[0.0em]{\textbf{Method}} & \multirow{1}{*}[0.0em]{\textbf{Decoder}} & \shortstack{\textbf{Training} \\ (ms/step)} & \shortstack{\textbf{Inference} \\ (ms/image)}  \\ \midrule
MNV2 + ASPP & - & 251.79 & 11.20 \\
LI-MNV2 + LI-ASPP & - & 252.62 & 11.26 \\
\midrule
ResNet50 + ASPP & \checkmark & 534.90 & 20.66 \\
LI-ResNet50 + LI-ASPP & \checkmark & 535.40 & 20.73 \\
\bottomrule
\end{tabularx}
\end{table}

\textbf{Multi-scale Evaluations} \hspace{0.1cm} We further compare the performance of different methods when applying the multi-scale evaluation techniques \cite{chen2017rethinking,deeplabv3plus2018}. 
Particularly, we evaluate results of the LI-based models and the baseline on Pascal Voc 2012 validation set using three different multi-scale settings and with Deeplabv3+ Decoder \cite{deeplabv3plus2018} disabled/enabled. As shown in Table \ref{tab: performance_multi_eval_scales}, the application of the multi-scale techniques significantly increases the segmentation accuracy of both our and baseline models, while our method consistently outperforms the baseline no matter which multi-scale setting is employed. This is following our expectations, since the proposed LI-Convs can fundamentally enhance the model's sensitivity to semantic contours, thus will benefit the segmentation results of varying input scales. Additionally, our LI models can work seamlessly with Deeplabv3+ Decoder to achieve the highest mIoUs for all multi-scale settings, which again verifies the generality of LI-Convs.

\begin{figure*}[ht!]
\centering
\includegraphics[width=0.9\linewidth]{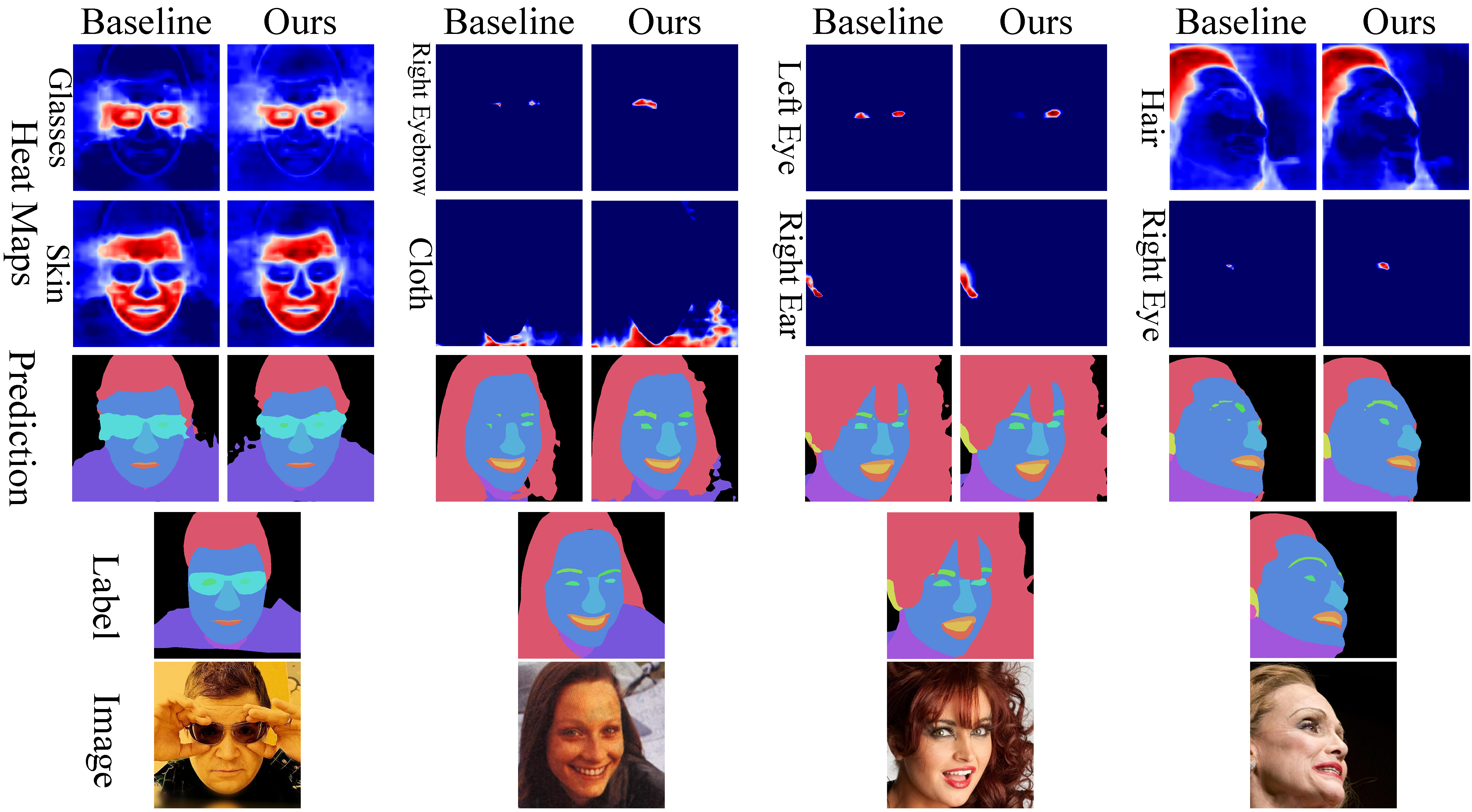}
\caption{Visualisations of the class-level heat maps and semantic predictions of the baseline (MNV2+ASPP) and our method (LI-MNV2+LI-ASPP) on CelebAMask-HQ. Deeper reds in heat maps represent higher positive responses or more attention from the model, and vice versa for deeper blues. Our method allocates more attention to shape the semantic boundary areas and thus can produce predictions with higher visual qualities.}
\label{fig: heatmap}
\end{figure*}

\begin{figure*}[t!]
\centering
\includegraphics[width=0.9\linewidth]{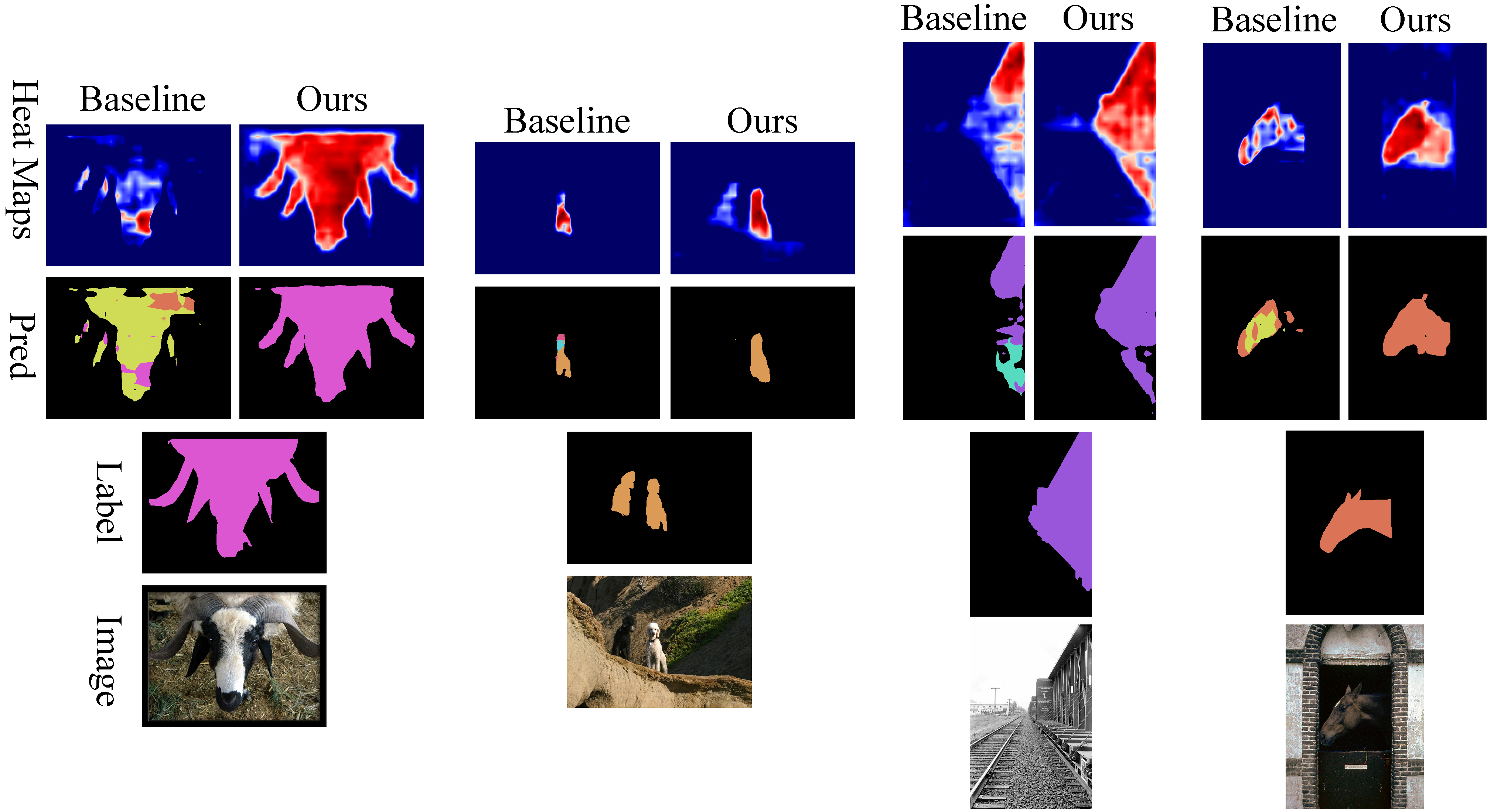}
\caption{Visualisations of the class-level heat maps and semantic predictions of the baseline (MNV2+ASPP) and our method (LI-MNV2+LI-ASPP) on Pascal Voc 2012 dataset. Deeper reds in heat maps represent higher positive responses or more attention from the model, and vice versa for deeper blues.}
\label{fig: heatmap_pascalVoc2012}
\end{figure*}

\textbf{Running Speed} We also compare the actual running speed before and after adding the LI layers. 
Without loss of generality, Table \ref{tab: speed_pascal} shows the actual training and inference time of two baselines and our LI-based models on Pascal Voc 2012 with RTX-2080Ti GPU. The decoder \cite{deeplabv3plus2018} is disabled for MNV2 backbone and is enabled for ResNet-50 backbone to represent two different scenarios, i.e. efficiency or performance first. As can be seen from the table, the LI layers only add less than 1 ms (0.1\%-0.33\% increase) to the training time at each step, and less than 0.1 ms (0.33\%-0.55\% increase) to the inference time. Considering the significant performance improvement, our LI layer is arguably a light-weight module that can be easily integrated into deep segmentation models to enhance their segmentation powers.

\subsection{Discussion}

\begin{figure*}[t!]
\centering
\includegraphics[width=0.9\linewidth]{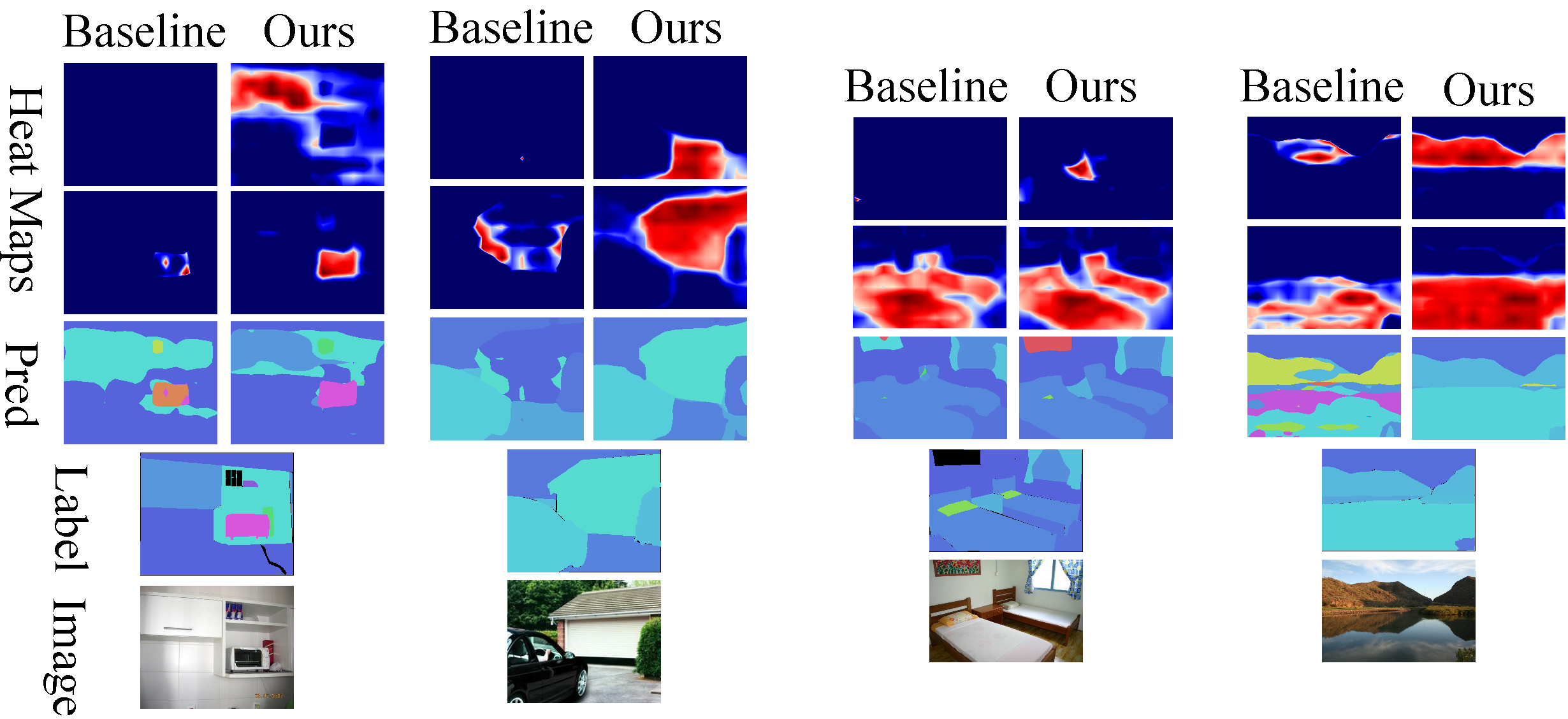}
\caption{Visualisations of the class-level heat maps and semantic predictions of the baseline (MNV2+ASPP) and our method (LI-MNV2+LI-ASPP) on ADE20K dataset. Deeper reds in heat maps represent higher positive responses or more attention from the model, and vice versa for deeper blues.}
\label{fig: heatmap_ADE20K}
\end{figure*}

\textbf{How the LI layer works} \hspace{0.1cm} To intuitively understand the LI mechanisms, we dive into the channel-level features to visualise the patterns before and after LI layers. 
As demonstrated in Fig. \ref{fig: channel_LI_plots}, we plot several feature channels before and after the LI layers in LI-ASPP on CelebAMask-HQ dataset. It can be discovered that although the intensity of activation is suppressed globally after the LI layer, the inhibited feature exhibits more recognisable patterns with clarified and emphasised contours, which can be more desirable in the segmentation domain.

\textbf{What interests the model} \hspace{0.1cm} In Fig. \ref{fig: heatmap}, we visualise the class-level heat maps and the segmentation predictions generated by the baseline (MNV2+ASPP) and our method (LI-MNV2+LI-ASPP) on CelebAMask-HQ. We utilise deeper reds to denote higher positive neurons responses (more model attention) in heat maps, and vice versa for deeper blues. Compared with the baseline, the semantically meaningful contouring areas receive more attention from our model, e.g. the \say{glasses} and \say{skin} heat maps in Fig. \ref{fig: heatmap}. Such kind of contour sensitivity can be reasonably attributed to the proposed LI-Convs. Besides, the segmentation predictions generated by our method have better visual qualities, which also verifies the superiority of the LI-Convs.   

In Fig. \ref{fig: heatmap_pascalVoc2012} and Fig. \ref{fig: heatmap_ADE20K}, the predictions and heatmaps on Pascal Voc 2012 and ADE20K datasets are illustrated to provide comparisons between the baseline (MNV2+ASPP) result and that of the proposed method (LI-MNV2+LI-ASPP). It can be seen that the heatmaps from our method generally captures more accurate semantic areas than the baseline on both datasets, and thus the segmentation labels generated by our method are of higher visual qualities. Since the improvement on contour has been evident on all three datasets, we further demonstrate the effectiveness and generality of the proposed LI-Convs.
\section{Conclusion}
We describe a dilated convolution with lateral inhibitions (LI-Convs) to enhance the model's sensitivity to semantic contours and to extract features at denser scales. The performance of the proposed LI-ASPP, LI-MNV2 and LI-ResNet architectures is shown to outperform the baseline method on three segmentation benchmark datasets, which verify the effectiveness and generality of the LI-Convs. We also investigate and try to understand the working mechanisms hidden behind. The proposed LI-Convs can be seamlessly integrated into deep models for other tasks, such as lip-reading and object detection, that require explicit awareness of the semantic boundaries.

\bibliographystyle{IEEEtran}
\bibliography{ref.bib}

\end{document}